\documentclass[conference]{IEEEtran}
\IEEEoverridecommandlockouts
\usepackage{cite}
\usepackage{amsmath,amssymb,amsfonts}
\usepackage{algorithmic}
\usepackage{graphicx}
\usepackage{textcomp}
\usepackage{xcolor}
\usepackage{url}
\usepackage[top=0.75in, bottom=1.05in, left=0.625in, right=0.625in]{geometry}
\def\BibTeX{{\rm B\kern-.05em{\sc i\kern-.025em b}\kern-.08em
    T\kern-.1667em\lower.7ex\hbox{E}\kern-.125emX}}
\begin{document}

\title{Synthetic Data Augmentation for Enhancing Harmful Algal Bloom Detection with Machine Learning}

\author{\IEEEauthorblockN{Tianyi Huang}
\IEEEauthorblockA{\textit{Student Ambassador} \\
\textit{App Inventor Foundation}\\
Fremont, California, USA \\
tianyi@appinventorfoundation.org}
}

\IEEEpubid{\begin{minipage}{\textwidth}\ \\[12pt]
  Accepted for publication at the 2025 IEEE Conference on Technologies for \\ Sustainability (SusTech)
\end{minipage}} 

\maketitle

\begin{abstract}

Harmful Algal Blooms (HABs) pose severe threats to aquatic ecosystems and public health, resulting in substantial economic losses globally. Early detection is crucial but often hindered by the scarcity of high-quality datasets necessary for training reliable machine learning (ML) models. This study investigates the use of synthetic data augmentation using Gaussian Copulas to enhance ML-based HAB detection systems. Synthetic datasets of varying sizes (100--1,000 samples) were generated using relevant environmental features---water temperature, salinity, and UVB radiation---with corrected Chlorophyll-\textit{a} concentration as the target variable. Experimental results demonstrate that moderate synthetic augmentation significantly improves model performance (RMSE reduced from 0.4706 to 0.1850; \( p < 0.001 \)). However, excessive synthetic data introduces noise and reduces predictive accuracy, emphasizing the need for a balanced approach to data augmentation. These findings highlight the potential of synthetic data to enhance HAB monitoring systems, offering a scalable and cost-effective method for early detection and mitigation of ecological and public health risks.

\end{abstract}


\section{Introduction}
Harmful Algal Blooms (HABs) are ecological phenomena characterized by the rapid proliferation of algae in aquatic environments. They are caused by multiple factors, including excessive nutrient loads, rising water temperatures, and other anthropogenic influences \cite{anderson2002, paerl2013}. HABs pose significant threats to aquatic ecosystems, public health, and economies worldwide \cite{wurtsbaugh2019}. In the United States alone, these events are estimated to cause \$50 million in annual economic damages, affecting sectors such as fisheries, tourism, and public water supplies \cite{dobricic2016, hoagland2002}. A notable example is the 2014 Lake Erie water crisis, which deprived over 400,000 residents of safe drinking water, highlighting the urgent need for efficient and accurate HAB detection systems \cite{steffen2014}.

Current HAB monitoring methods rely heavily on manual sampling and laboratory analysis, both of which are time-intensive and costly \cite{wells2015}. As a result, these traditional approaches fall short of the real-time detection necessary to minimize the adverse impacts of HABs. Machine learning (ML) offers a promising pathway for early HAB detection by using environmental data to predict algal bloom risks \cite{hill2019}. However, the lack of high-quality and balanced datasets—particularly for extreme events—poses a major hurdle to training robust ML models \cite{bamra2022}.

To mitigate the issue of data scarcity, synthetic data generation has emerged as a practical solution. Synthetic datasets can complement existing data by filling gaps in underrepresented regions of the data distribution \cite{molares2024, toutouh2020}. While prior efforts have primarily explored synthetic data generation using Large Language Models (LLMs), this study employs Gaussian Copulas, a probabilistic method known for preserving complex interdependencies within data \cite{nelsen2007}. This study aims to systematically investigate how varying volumes of synthetic data generated via Gaussian Copulas influence the performance of machine learning models designed for HAB detection.

\section{Related Works}
In recent years, the application of machine learning (ML) techniques for the detection and prediction of Harmful Algal Blooms (HABs) has been extensively investigated. For example, Hill et al. developed HABNet, a framework integrating remote sensing data with ML models such as Convolutional Neural Networks (CNNs) and Long Short-Term Memory (LSTM) networks to detect and predict HAB events \cite{hill2019}. Their study demonstrated the viability of using ML models to process complex spatiotemporal data for environmental monitoring. Likewise, Molares-Ulloa et al. examined hybrid ML methodologies for mitigating the impacts of HABs \cite{molares2024}. By comparing models like Neural-Network-Adding Bootstrap (BAGNET) and Discriminative Nearest Neighbor Classification (SVM-KNN) in estimating the state of production areas affected by algal blooms, they showcased the effectiveness of hybrid approaches in detecting bloom episodes across different estuaries.

In the realm of synthetic data generation, Toutouh employed Conditional Generative Adversarial Networks (GANs) to model urban outdoor air pollution, aiming to create realistic pollution time-series data for environmental monitoring \cite{toutouh2020}. Similarly, Wang et al. introduced a feature-supervised GAN for environmental monitoring during hazy days, illustrating the usefulness of synthetic data in bolstering model performance under challenging visibility conditions \cite{wang2020}.

Expanding on more recent studies that uses synthetic data augmentation in ecological and environmental contexts, Sousa et al. proposed a diffusion model approach for data augmentation in Earth observation, addressing issues of data scarcity and diversity in satellite imagery \cite{Sousa2024}. Their method leverages meta-prompts for instruction generation, vision-language models to create captions, and fine-tunes an Earth Observation diffusion model to iteratively enhance data diversity. On the other hand, Glazkova et al. employed prompt-based data augmentation strategies using LLMs to improve multi-label text classification in ecological texts \cite{Glazkova2024}. By testing different prompt formulations—ranging from paraphrasing existing texts to generating entirely new samples—the authors observed substantial improvements in classification performance, demonstrating the efficacy of synthetic text augmentation for ecological research. Collectively, these studies reveal the growing importance of augmentation techniques that address the persistent constraints of data scarcity and imbalance across environmental domains.

While these studies illustrate the potential of ML-driven and synthetic data approaches for environmental monitoring, there remains a gap in understanding how the volume of synthetic data affects model performance, as well as the efficacy of Gaussian Copulas for synthetic data generation in ecological contexts—an approach that can be significantly more cost-effective than many alternatives. In response, this study implements a complete ML pipeline that leverages Gaussian Copulas to generate synthetic data for HAB detection, rigorously assessing how varying volumes of this synthetic data affect both model accuracy and generalization capabilities. This exploration offers a novel perspective on mitigating data scarcity and imbalance in environmental datasets, ultimately contributing to the development of more effective and reliable predictive models for monitoring HABs and other environmental phenomena.

\section{Methodology}

\begin{figure}[ht]
\centerline{\includegraphics[width=\columnwidth]{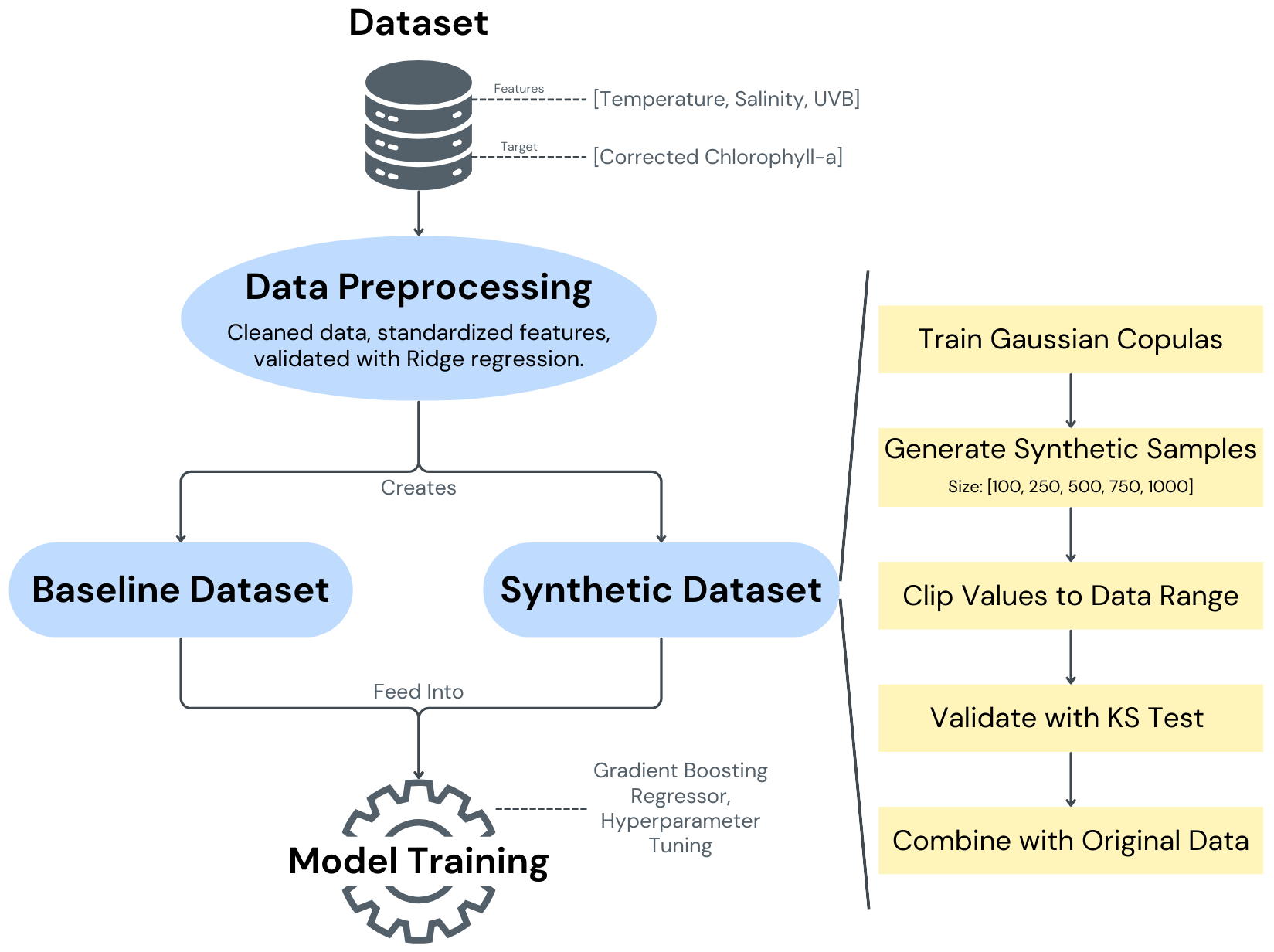}}
\caption{
Illustration of Methodology Pipeline. (1) \textit{Data Preprocessing}, where raw measurements are cleaned, imputed, and standardized, followed by baseline validation using Ridge regression; (2) \textit{Synthetic Data Generation}, in which Gaussian Copulas learn the joint distribution of real measurements, generate additional samples, and undergo Kolmogorov--Smirnov testing to ensure distributional similarity; (3) \textit{Model Training}, where both the baseline dataset and synthetic-augmented dataset are used to train Gradient Boosting Regressors with hyperparameter tuning. Each step is designed to systematically evaluate the impact of synthetic data on predictive performance.
}
\label{fig:method}
\end{figure}

\subsection{Data Collection}
This study used the \textit{Dataset of the high-frequency biological, meteorological, and hydrological parameters and weekly water sampling}, specifically the "High Freq 2015" subset from the study \textit{Water temperature drives phytoplankton blooms in coastal waters}~\cite{trombetta2019}. The original dataset included a wide range of variables, such as temperature, salinity, chlorophyll-\textit{a} fluorescence, turbidity, wind speed, and more. For the purpose of this study, key parameters—water temperature (\textdegree{}C), salinity (PSU), and UVB radiation (mW/m\textsuperscript{2})—were selected due to their relevance to harmful algal bloom (HAB) prediction. The dataset consists of 12,657 rows of measurements, with the corrected Chlorophyll-\textit{a} concentration (\textmu{}g/L) chosen as the target variable because of its established significance as an indicator of HAB events.

\subsection{Data Preprocessing}
The preprocessing pipeline for the study's baseline (non-synthetic) dataset involved several key steps to prepare the raw data for machine learning models. Missing values were removed during the cleaning phase to maintain data integrity, and median imputation was applied--a technique where missing values are replaced with the median of the corresponding feature--to fill any gaps in the features. Median imputation has been shown to maintain statistical robustness, particularly in datasets with environmental variability.

Once the data was cleaned, it was split into training and testing sets using an 80:20 ratio. This ensured that model evaluation was based on unseen data, reducing bias in performance assessment. Features were standardized using the StandardScaler function from scikit-learn, which transforms each feature so that it has zero mean and unit variance, normalizing the range of input variables \cite{pedregosa2011}. This ensures that features contribute equally to the model training process, an important step for effective convergence in ML algorithms. This linear regression model with L2 regularization penalizes large coefficient values to reduce overfitting and provides a baseline to assess the predictive capacity of the dataset, serving as a benchmark for further experimentation with synthetic data. Ridge regression is particularly suitable for multicollinearity datasets or when predictive accuracy is prioritized \cite{hoerl2000}.

\subsection{Synthetic Data Generation and Integration}
To address the challenges of data scarcity and imbalance, synthetic data was generated using Gaussian Copulas, a probabilistic method known for its ability to capture and replicate the multivariate dependencies present in the original dataset. Gaussian Copulas have been widely used in fields such as finance and environmental modeling to generate realistic synthetic data \cite{nelsen2007}. 

\textbf{Configuration and Assumptions.} In this study, the \textit{GaussianMultivariate} class from the \textit{copulas} library was employed with its default settings, which automatically fit univariate distributions for each variable and then modeled the correlation structure among them. No additional parameter constraints were manually imposed, allowing the copula model to learn the joint distribution based on the observed data. As a key assumption, the variables were treated as continuous, and any out-of-range synthetic values were clipped to the minimum or maximum of the corresponding real-data feature. This approach preserves data realism and guards against introducing extreme outliers.

\textbf{Synthetic Sample Generation.} The cleaned and transformed data served as the basis for training the Gaussian Copula model, which learned the joint distribution of the features and the target variable. Synthetic samples were generated at varying volumes—100, 250, 500, 750, and 1000 rows—to systematically evaluate the impact of synthetic data augmentation on model performance. The generated synthetic data was clipped to the range of the original dataset to ensure realistic values and prevent the introduction of noise or outliers. This step was critical for maintaining data validity and minimizing the risk of misleading results.

\textbf{Distribution Validation.} A Kolmogorov-Smirnov (KS) test was conducted to validate the similarity between the distributions of real and synthetic data. The KS test, a non-parametric method for comparing two distributions, confirmed that the synthetic samples retained the statistical properties of the original dataset \cite{massey1951}.

\textbf{Data Integration.} The synthetic data was then combined with the original dataset in varying proportions to test its influence on predictive accuracy. To capture complex interactions between variables, polynomial feature expansion was applied, generating higher-order features. The combined datasets were subjected to the same preprocessing steps as the baseline dataset, including splitting into training and testing sets, imputing missing values, and standardizing features.

\subsection{Model Training}
The Gradient Boosting Regressor, implemented using scikit-learn, was selected as the primary model to analyze the effect of synthetic data on predictive performance \cite{pedregosa2011}. Gradient Boosting is particularly effective at capturing non-linear relationships and handling structured data, making it a strong candidate for this study’s predictive task \cite{friedman2001}. Separate experiments were conducted using synthetic-augmented datasets and non-synthetic datasets to isolate the impact of synthetic data on the training process.

\textbf{Hyperparameter Tuning.} To ensure optimal performance, hyperparameter tuning was applied using GridSearchCV, a widely adopted tool for the systematic exploration of parameter spaces \cite{pedregosa2011}. In particular, the search spanned key Gradient Boosting parameters:
\begin{itemize}
    \item \textit{n\_estimators}: [100, 200, 300]
    \item \textit{learning\_rate}: [0.01, 0.05, 0.1]
    \item \textit{max\_depth}: [3, 5, 7]
\end{itemize}
Each combination of parameters was evaluated under 5-fold cross-validation to balance computational efficiency and thoroughness. The final model selection was guided by the highest average \(R^2\) score across folds. This approach enabled the identification of configurations that best fit the training data without overfitting, ensuring a model that generalizes effectively.

\subsection{Experimental Setup}
To systematically evaluate the impact of synthetic data on the predictive accuracy of Gradient Boosting models for corrected Chlorophyll-\textit{a} concentrations, synthetic data was incrementally added to the original dataset in five levels: 100, 250, 500, 750, and 1000 samples. These levels were chosen to represent synthetic contributions below 10\% of the total dataset, ensuring that real data remained the dominant contributor to training.

In addition to models trained on synthetic-augmented datasets, a baseline model trained exclusively on real data was included for comparison. The experimental framework evaluated each model’s performance using the following metrics:
\begin{itemize}
    \item \textbf{Mean Squared Error (MSE):} Quantifies the average squared difference between predicted and actual values, penalizing larger errors more heavily.
    \item \textbf{Root Mean Squared Error (RMSE):} Provides a more interpretable measure by taking the square root of MSE, expressed in the same units as the target variable \cite{willmott2005}.
    \item \textbf{Mean Absolute Error (MAE):} Measures the average magnitude of errors without considering their direction, offering insights into typical prediction deviations.
    \item \textbf{Percent Error:} Evaluates the relative accuracy of predictions, expressed as the percentage deviation from true values.
\end{itemize}

By comparing metrics across different synthetic data levels and the baseline, this setup provided insights into the role of synthetic data in improving model predictions.

\section{Results}

\subsection{Cross-Validation and Evaluation Metrics}
The cross-validation results show significant differences in performance across models trained on real data versus those augmented with synthetic data generated using Gaussian Copulas. Table I summarizes the Mean Squared Error (MSE), Root Mean Squared Error (RMSE), and Mean Absolute Error (MAE) for the baseline and synthetic-augmented models.

\begin{table}[htbp]
\caption{Performance Metrics Across Models}
\begin{center}
\begin{tabular}{|l|c|c|c|}
\hline
\textbf{Model} & \textbf{MSE} & \textbf{RMSE} & \textbf{MAE} \\
\hline
Base Model & 0.2215 & 0.4706 & 0.3499 \\
\hline
Synthetic 100 & 0.0352 & 0.1875 & 0.1452 \\
\hline
Synthetic 250 & 0.0342 & 0.1850 & 0.1464 \\
\hline
Synthetic 500 & 0.0377 & 0.1942 & 0.1529 \\
\hline
Synthetic 750 & 0.0389 & 0.1973 & 0.1541 \\
\hline
Synthetic 1000 & 0.0556 & 0.2358 & 0.1866 \\
\hline
\end{tabular}
\label{tab:metrics}
\end{center}
\end{table}

The baseline model exhibits the worst performance across all metrics, reflecting its inability to generalize effectively due to data scarcity. Models trained with 100 and 250 synthetic rows achieve a six-fold reduction in MSE compared to the baseline, demonstrating significant improvements in predictive accuracy. However, increasing the volume of synthetic data beyond 250 rows results in diminishing returns, with the model trained on 1000 synthetic rows performing worse than those trained on 100 or 250 rows. This trend suggests that excessive synthetic data may introduce noise or lead to overfitting.

\begin{figure}[h]
\centerline{\includegraphics[width=\columnwidth]{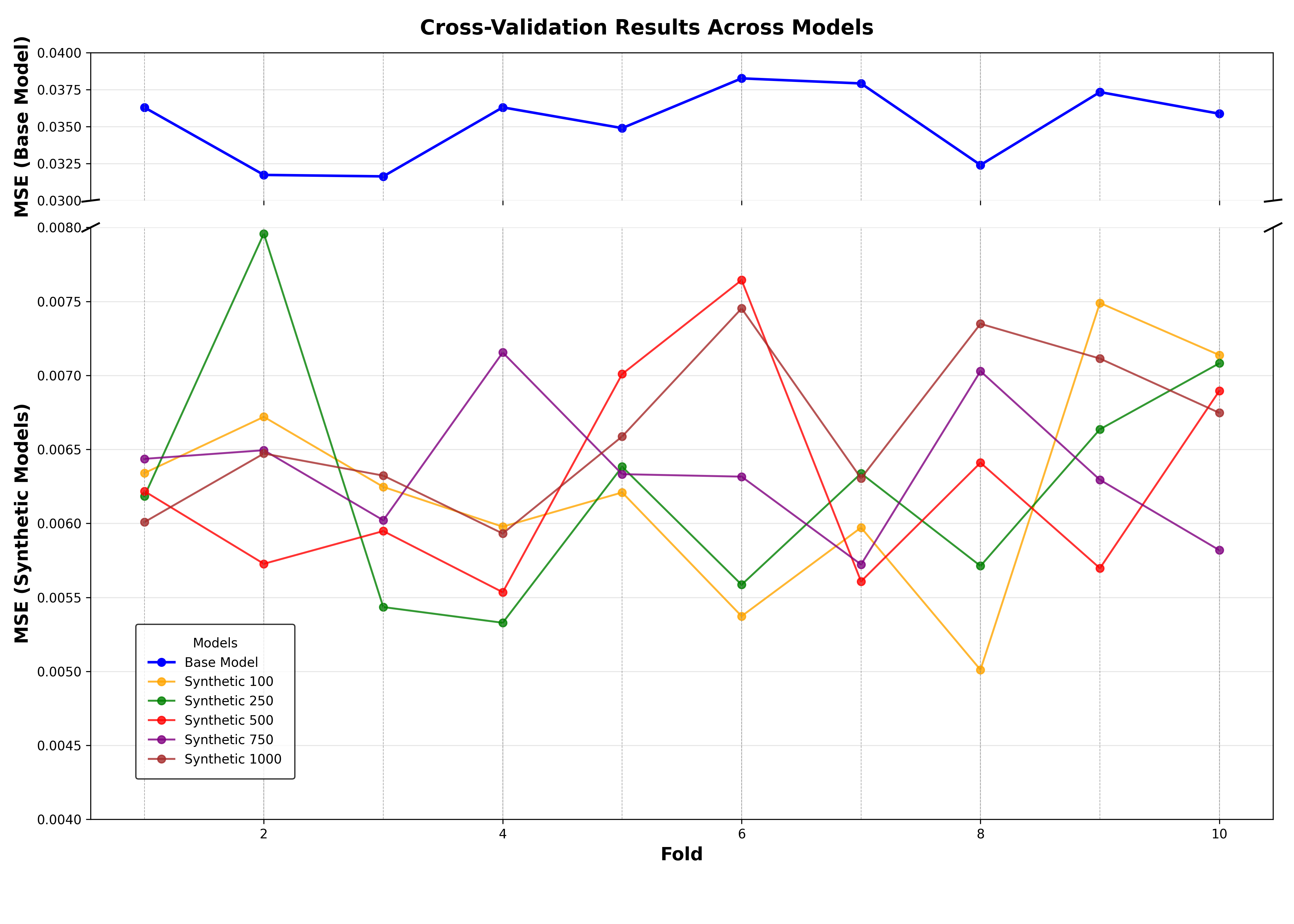}}
\caption{Cross-Validation Results Across Models. This plot shows the Mean Squared Error (MSE) over ten cross-validation folds for both the baseline model (trained on real data) and models augmented with varying amounts of synthetic data.}
\label{fig:cv_results}
\end{figure}

Figure 2 plots the cross-validation MSE for each fold, providing a visual comparison of model performance. The baseline model consistently exhibits the highest MSE across all folds, whereas synthetic models, particularly those with 100 and 250 rows, demonstrate lower and more stable MSE values.

\subsection{Percent Error Analysis}
Percent error analysis reveals additional insights into the predictive accuracy and variability of the models. Table II provides detailed statistics on the mean, median, and standard deviation of percent errors for each model.

\begin{table}[htbp]
\caption{Percent Error Statistics Across Models}
\begin{center}
\begin{tabular}{|l|c|c|c|}
\hline
\textbf{Model} & \textbf{Mean (\%)} & \textbf{Median (\%)} & \textbf{Std Dev (\%)} \\
\hline
Base Model & 10.17 & 7.73 & 9.09 \\
\hline
Synthetic 100 & 7.16 & 5.59 & 6.22 \\
\hline
Synthetic 250 & 7.21 & 5.68 & 6.17 \\
\hline
Synthetic 500 & 10.47 & 7.10 & 10.49 \\
\hline
Synthetic 750 & 7.71 & 6.09 & 6.62 \\
\hline
Synthetic 1000 & 7.11 & 5.46 & 6.40 \\
\hline
\end{tabular}
\label{tab:percent_error}
\end{center}
\end{table}

The baseline model has the highest mean percent error (10.17\%), indicating relatively poor prediction accuracy. Models trained with 100 and 250 synthetic rows achieve the lowest mean percent errors (7.16\% and 7.21\%, respectively), reflecting improved prediction reliability. Interestingly, the models trained on 750 and 1000 synthetic rows exhibit comparable mean percent errors (7.71\% and 7.11\%, respectively) but display greater variability, as indicated by their standard deviations. This suggests that while higher volumes of synthetic data can occasionally improve predictions, they also increase the likelihood of large deviations.

\begin{figure}[h]
\centerline{\includegraphics[width=\columnwidth]{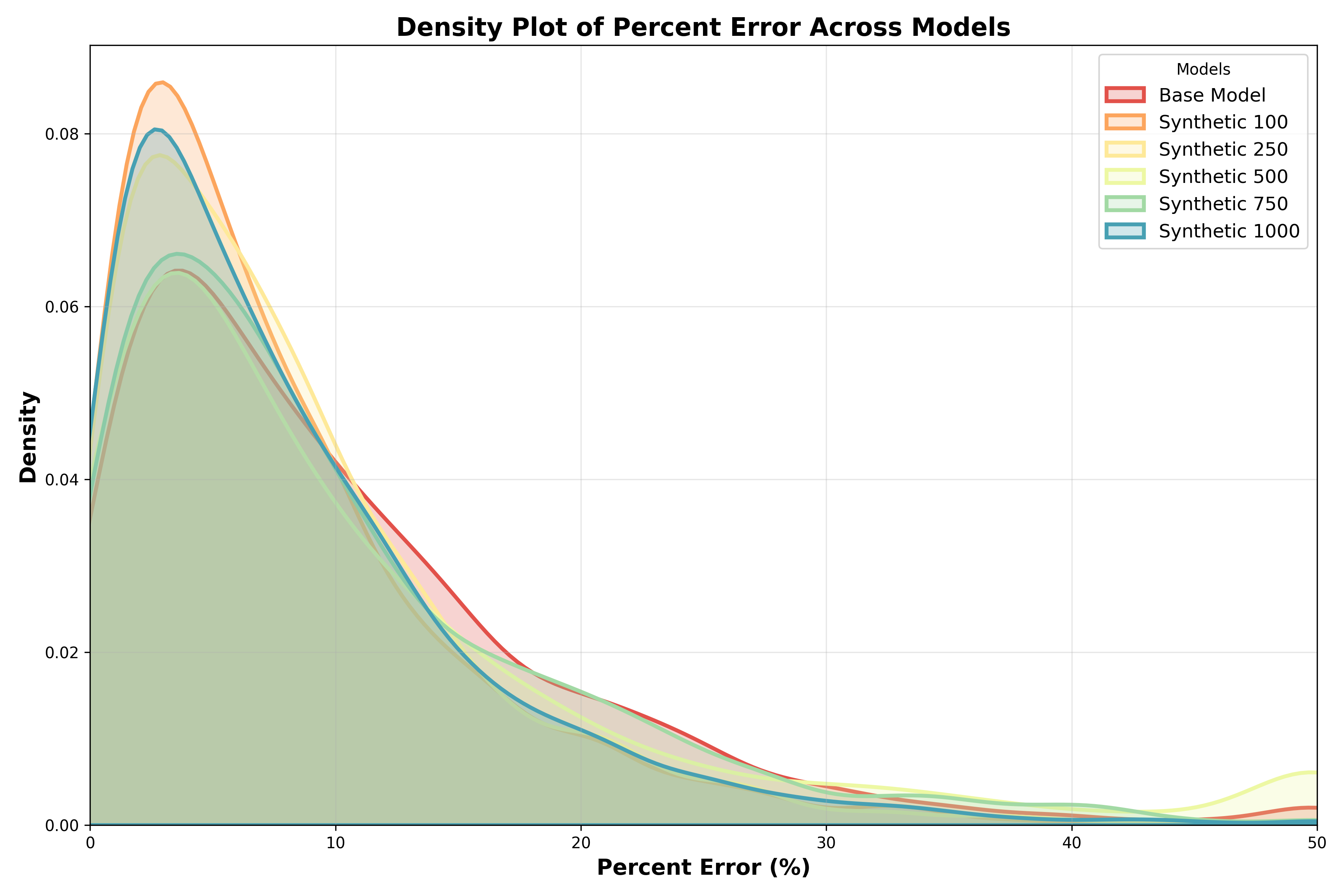}}
\caption{Density Plot of Percent Error Across Models. The distribution of percent errors is depicted for the baseline model and synthetic-augmented models.}
\label{fig:percent_error}
\end{figure}

Figure 3 illustrates the density distribution of percent errors for all models. The baseline model has a broader and flatter distribution, indicative of less reliable predictions. In contrast, synthetic models, particularly those with 100 and 250 rows, exhibit narrower distributions concentrated around lower percent error values, further confirming their superior accuracy.

\subsection{Statistical Significance Testing}
To assess the significance of performance improvements, paired t-tests were conducted between the baseline model and each synthetic model. The null hypothesis assumed no difference in performance between the models, while the alternative hypothesis posited that synthetic models outperform the baseline. The p-values for these comparisons are summarized in Table III.

\begin{table}[htbp]
\caption{P-Values for Paired T-Tests Comparing Baseline and Synthetic Models}
\begin{center}
\begin{tabular}{|l|c|c|}
\hline
\textbf{Comparison} & \textbf{p-value} & \textbf{Significant?} \\
\hline
Baseline vs. Synthetic 100 & $<$ 0.001 & Yes \\
\hline
Baseline vs. Synthetic 250 & $<$ 0.001 & Yes \\
\hline
Baseline vs. Synthetic 500 & 0.085 & No \\
\hline
Baseline vs. Synthetic 750 & 0.098 & No \\
\hline
Baseline vs. Synthetic 1000 & 0.231 & No \\
\hline
\end{tabular}
\label{tab:p_values}
\end{center}
\end{table}

The results indicate that the improvements achieved with 100 and 250 synthetic rows are statistically significant (\textit{p} $<$ 0.001). Conversely, models trained on 500 or more synthetic rows do not exhibit statistically significant improvements over the baseline, likely due to overfitting or noise in the training process.

\subsection{Comparative Advantages Over Existing Approaches}
Beyond the measurable gains in predictive accuracy, this Gaussian Copula--based framework offers several benefits that address known limitations in prior work:

\begin{itemize}
    \item \textbf{Preserving Interdependencies.} Unlike simple oversampling or LLM--based approaches that often fail to maintain essential correlations among variables (e.g., temperature, salinity, and UVB radiation) \cite{ding2024}, Gaussian Copulas effectively capture and retain these complex relationships. Moreover, LLM-generated data can lead to out-of-range values, hallucinations, and difficulties in ensuring structured outputs. By contrast, our copula-based method offers a more reliable way to integrate synthetic samples without additional complexity in adjusting output data.
    \item \textbf{Distributional Realism.} While some deep generative algorithms (e.g., GAN variants) are prone to mode collapse or excessive variance \cite{wiatrak2019}, Gaussian Copulas avoid these pitfalls by fitting explicit probability distributions and clipping synthetic values to realistic ranges. This process ensures that augmented data remains representative of the original variance and does not introduce outliers that could worsen model performance.
    \item \textbf{Computational Requirements.} Compared to advanced deep learning generators that demand extensive computational resources, the copula-based pipeline produces high-fidelity synthetic data with significantly fewer hardware requirements. For instance, in this study, we used a 12,657-row HAB dataset, and all computations ran reliably on a standard MacBook Air with an Apple M3 chip and 24 GB of memory, highlighting the practical feasibility of this approach for small- to medium-sized datasets.
\end{itemize}

\subsection{Implications of Findings} 
The findings of this study demonstrate the practicality of synthetic data augmentation for improving HAB detection and other environmental monitoring tasks, especially when moderate volumes (e.g., 100–250 rows) of Gaussian Copula–based synthetic samples are used. By effectively filling data gaps without introducing severe noise, this approach can strengthen early warning systems, reduce reliance on expensive and time-consuming sampling, and improve policy-driven interventions in vulnerable areas. Moreover, the clear demonstration of diminishing returns at higher synthetic volumes provides insight for balancing real and synthetic data. Looking ahead, integrating this framework with real-time sensor networks and more advanced machine learning architectures could further refine model performance and scalability, ultimately supporting more proactive and cost-effective management of environmental health risks.

\section{Limitations and Future Work}

\subsection{Limitations}
Although this study confirms the value of Gaussian Copula--based synthetic data for HAB detection, several practical constraints should be acknowledged:
\begin{itemize}
    \item \textbf{Dataset Diversity:} The dataset’s limited geographic scope and lack of nutrient-related features can restrict the model’s generalizability. Broader data collection efforts would be needed to ensure that the method scales effectively to different aquatic systems and bloom conditions.
    \item \textbf{Computational Overhead:} As datasets grow in size or complexity, fitting and sampling from Gaussian Copulas can become increasingly difficult. In high-volume or streaming scenarios, training times may increase sharply, potentially necessitating parallel computation or incremental updating strategies.
    \item \textbf{Resource Constraints:} This work did not explore advanced weighting schemes for synthetic data or deploy more sophisticated synthetic generators (e.g., GANs) due to time and computational limitations \cite{goodfellow2014}. Such restrictions also limited experimentation with a wider range of hyperparameter configurations.
    \item \textbf{Real-Time Deployment Gaps:} The present approach has only been validated with historical data. Transitioning to a real-time monitoring framework would require sensor networks, automated data ingestion pipelines, and consistent checks to maintain data quality and model accuracy.
\end{itemize}
\subsection{Future Work}
Several avenues can further strengthen and extend the findings presented here:
\begin{itemize}
    \item \textbf{Advanced Machine Learning Algorithms:} A structured evaluation of models like XGBoost and LightGBM may reveal additional performance gains and elucidate how these algorithms handle synthetic datasets \cite{friedman2001, chen2016}. A potential roadmap includes: (1) conducting initial comparisons of these models with Gradient Boosting to assess baseline improvements in predictive metrics, (2) applying Bayesian optimization to fine-tune key hyperparameters (e.g., learning rate, maximum tree depth, etc.), and (3) evaluating their ability to handle synthetic data with varying proportions to identify optimal augmentation strategies.
    \item \textbf{Geographic and Feature Range:} Incorporating data from diverse regions and additional variables such as nutrient profiles could enhance the representativeness of copula-based synthetic augmentation, improving the reliability of HAB forecasts.
    \item \textbf{Real-Time Integration:} Implementing low-cost, in-situ sensors for continuous data collection would enable live updates to the training pipeline. This approach could allow incremental or online learning, aligning HAB detection systems more closely with real-world monitoring needs.
    \item \textbf{Alternative Synthetic Methods:} Experimenting with other generative techniques—such as GANs and Variational Autoencoders—could potentially further refine synthetic data quality, particularly for larger and more heterogeneous datasets \cite{kingma2013, goodfellow2014}.
\end{itemize}

\section{Conclusion}
This study successfully demonstrates the potential of synthetic data generated via Gaussian Copulas to improve predictive modeling for Harmful Algal Bloom (HAB) detection. In particular, moderate synthetic volumes (100--250 rows) delivered the most significant improvements, with the best-performing model reaching a 7.21\% mean percent error—a 29.09\% improvement over the 10.17\% error observed in the baseline. These results showcase the effectiveness of synthetic data in addressing data scarcity and enhancing model accuracy. However, the results also emphasize the importance of a balanced approach, as excessive augmentation (e.g., 1000 rows) introduced noise and reduced predictive accuracy. 

Beyond confirming the feasibility of copula-based synthetic data in improving HAB forecasts, this work accentuates the broader need for collaborative efforts among researchers, policymakers, and environmental agencies. By integrating synthetic augmentation into larger monitoring frameworks and real-time sensor networks, stakeholders can develop more proactive mitigation strategies for at-risk aquatic ecosystems. Adopting these techniques could potentially expedite data-driven policy decisions, strengthen conservation measures, and reduce the ecological and economic toll of HABs on communities worldwide.

\bibliographystyle{IEEEtran}

\begin{thebibliography}{00}

\bibitem{anderson2002} D. M. Anderson, P. M. Glibert, and J. M. Burkholder, “Harmful algal blooms and eutrophication: Nutrient sources, composition, and consequences,” Estuaries, vol. 25, no. 4, pp. 704–726, Aug. 2002, doi: https://doi.org/10.1007/bf02804901.

\bibitem{paerl2013} H. Paerl, “Mitigating Harmful Cyanobacterial Blooms in a Human- and Climatically-Impacted World,” Life, vol. 4, no. 4, pp. 988–1012, Dec. 2014, doi: https://doi.org/10.3390/life4040988.

\bibitem{wurtsbaugh2019} W. A. Wurtsbaugh, H. W. Paerl, and W. K. Dodds, “Nutrients, eutrophication and harmful algal blooms along the freshwater to marine continuum,” Wiley Interdisciplinary Reviews: Water, vol. 6, no. 5, Aug. 2019, doi: https://doi.org/10.1002/wat2.1373.

\bibitem{dobricic2016} Joint Research Centre (European Commission), S. Dobricic, L. Pozzoli, I. Sanseverino, D. Conduto, and T. Lettieri, Algal bloom and its economic impact. LU: Publications Office of the European Union, 2016. Available: https://op.europa.eu/en/publication-detail/-/publication/4d384d1b-1804-11e6-ba9a-01aa75ed71a1/language-en

\bibitem{hoagland2002} P. Hoagland, D. M. Anderson, Y. Kaoru, and A. W. White, “The economic effects of harmful algal blooms in the United States: Estimates, assessment issues, and information needs,” Estuaries, vol. 25, no. 4, pp. 819–837, Aug. 2002, doi: https://doi.org/10.1007/bf02804908.

\bibitem{steffen2014} M. M. Steffen, B. S. Belisle, S. B. Watson, G. L. Boyer, and S. W. Wilhelm, “Status, causes and controls of cyanobacterial blooms in Lake Erie,” Journal of Great Lakes Research, vol. 40, no. 2, pp. 215–225, Jun. 2014, doi: https://doi.org/10.1016/j.jglr.2013.12.012.

\bibitem{wells2015} M. L. Wells et al., “Harmful algal blooms and climate change: Learning from the past and present to forecast the future,” Harmful Algae, vol. 49, pp. 68–93, Nov. 2015, doi: https://doi.org/10.1016/j.hal.2015.07.009.

\bibitem{hill2019} H. P. R, A. Kumar, M. Temimi, and B. D. R, “HABNet: Machine Learning, Remote Sensing Based Detection and Prediction of Harmful Algal Blooms,” arXiv.org, 2019. https://arxiv.org/abs/1912.02305.

\bibitem{bamra2022} N. Bamra, V. Voleti, A. Wong, and J. Deglint, “Towards Generating Large Synthetic Phytoplankton Datasets for Efficient Monitoring of Harmful Algal Blooms,” arXiv (Cornell University), Jan. 2022, doi: https://doi.org/10.48550/arxiv.2208.02332.

\bibitem{molares2024} A. Molares-Ulloa, D. Rivero, J. G. Ruiz, E. Fernandez-Blanco, and L. de-la-Fuente-Valentín, “Hybrid machine learning techniques in the management of harmful algal blooms impact,” Computers and Electronics in Agriculture, vol. 211, p. 107988, Aug. 2023, doi: https://doi.org/10.1016/j.compag.2023.107988.

\bibitem{toutouh2020} J. Toutouh, “Conditional Generative Adversarial Networks to Model Urban Outdoor Air Pollution,” Communications in computer and information science, pp. 90–105, Jan. 2021, doi: https://doi.org/10.1007/978-3-030-69136-3\_7.

\bibitem{nelsen2007} R. B. Nelsen, “An Introduction to Copulas,” SpringerLink, 2021, doi: https://doi.org/10.1007-0-387-28678-0.

\bibitem{wang2020} K. Wang, S. Zhang, J. Chen, F. Ren, and L. Xiao, “A feature-supervised generative adversarial network for environmental monitoring during hazy days,” Science of The Total Environment, vol. 748, pp. 141445–141445, Dec. 2020, doi: https://doi.org/10.1016/j.scitotenv.2020.141445.

\bibitem{Sousa2024} T. Sousa, B. Ries, and N. Guelfi, “Data Augmentation in Earth Observation: A Diffusion Model Approach,” arXiv.org, 2024. https://arxiv.org/abs/2406.06218.

\bibitem{Glazkova2024} A. Glazkova and O. Zakharova, “Evaluating LLM Prompts for Data Augmentation in Multi-label Classification of Ecological Texts,” arXiv.org, 2024. https://arxiv.org/abs/2411.14896.

\bibitem{trombetta2019} T. Trombetta, F. Vidussi, S. Mas, D. Parin, M. Simier, and B. Mostajir, “Water temperature drives phytoplankton blooms in coastal waters,” PLOS ONE, vol. 14, no. 4, p. e0214933, Apr. 2019, doi: https://doi.org/10.1371/journal.pone.0214933.

\bibitem{pedregosa2011} F. Pedregosa et al., “Scikit-learn: Machine Learning in Python Pedregosa, Varoquaux, Gramfort et al,” Journal of Machine Learning Research, vol. 12, pp. 2825–2830, 2011, Available: https://arxiv.org/pdf/1201.0490

\bibitem{hoerl2000} A. E. Hoerl and R. W. Kennard, “Ridge Regression: Biased Estimation for Nonorthogonal Problems,” Technometrics, vol. 42, no. 1, p. 80, Feb. 2000, doi: https://doi.org/10.2307/1271436.

\bibitem{massey1951} F. J. Massey, “The Kolmogorov-Smirnov Test for Goodness of Fit,” Journal of the American Statistical Association, vol. 46, no. 253, pp. 68–78, 1951, doi: https://doi.org/10.2307/2280095.

\bibitem{friedman2001} J. H. Friedman, “Greedy function approximation: A gradient boosting machine,” The Annals of Statistics, vol. 29, no. 5, pp. 1189–1232, Oct. 2001, doi: https://doi.org/10.1214/aos/1013203451.

\bibitem{willmott2005} C. Willmott and K. Matsuura, “Advantages of the mean absolute error (MAE) over the root mean square error (RMSE) in assessing average model performance,” Climate Research, 2005, Available: https://www.semanticscholar.org/paper/Advantages-of-the-mean-absolute-error-(MAE)-over-in-Willmott-Matsuura/581d0024eccf55493dd7d63554063a683bef6103

\bibitem{ding2024} B. Ding et al., “Data Augmentation using LLMs: Data Perspectives, Learning Paradigms and Challenges,” Findings of the Association for Computational Linguistics ACL 2024, pp. 1679–1705, 2024, doi: https://doi.org/10.18653/v1/2024.findings-acl.97.

\bibitem{wiatrak2019} M. Wiatrak, S. V. Albrecht, and A. Nystrom, “Stabilizing Generative Adversarial Networks: A Survey,” arXiv.org, 2019. https://arxiv.org/abs/1910.00927.

\bibitem{goodfellow2014} I. J. Goodfellow et al., “Generative Adversarial Networks,” arXiv.org, Jun. 10, 2014. https://arxiv.org/abs/1406.2661

\bibitem{chen2016} T. Chen and C. Guestrin, “XGBoost: a Scalable Tree Boosting System,” Proceedings of the 22nd ACM SIGKDD International Conference on Knowledge Discovery and Data Mining - KDD ’16, pp. 785–794, 2016, doi: https://doi.org/10.1145/2939672.2939785.

\bibitem{kingma2013} D. P. Kingma and M. Welling, “Auto-Encoding Variational Bayes,” arXiv.org, Dec. 20, 2013. https://arxiv.org/abs/1312.6114

\bibitem{openai2024} OpenAI, “GPT-4o: A multimodal language model,” Openai.com, 2024. https://openai.com/index/hello-gpt-4o/

\end{thebibliography}

\appendices

\section{Code Availability}
All code for this research is available at \url{https://github.com/Tonyhrule/Synthetic-HAB-ML-Augmentation}

\end{document}